\documentclass[conference]{IEEEtran}
\IEEEoverridecommandlockouts
\usepackage{cite}
\usepackage{amsmath,amssymb,amsfonts}
\usepackage{algorithmic}
\usepackage{graphicx}
\usepackage{textcomp}
\usepackage{xcolor}
\def\BibTeX{{\rm B\kern-.05em{\sc i\kern-.025em b}\kern-.08em
    T\kern-.1667em\lower.7ex\hbox{E}\kern-.125emX}}

\usepackage[caption=false]{subfig}

\begin{document}

\title{Towards a Modular Bin-picking Framework for Handling Object Pose Uncertainties   
\thanks{The work was supported by Fabrikant Vilhelm Pedersen og Hustrus Legat.}
}

\author{\IEEEauthorblockN{1\textsuperscript{st} Frederik Hagelskjær*}
\IEEEauthorblockA{\textit{SDU Robotics} \\
\textit{The Mærsk Mc-Kinney Møller Institute} \\
\textit{University of Southern Denmark}\\
Odense, Denmark \\
frhag@mmmi.sdu.dk}
*Corresponding author
}

\maketitle

\begin{abstract}
In recent years, there has been growing interest in robust robotic systems for precise bin-picking applications. To achieve reliable performance, such systems must address errors arising from both the object pose estimation and the grasping process. Although various approaches have been proposed, they typically target specific challenges and do not offer general solutions. In this paper, we present a modular framework that jointly handles both error types. The framework incorporates object pose distribution estimation to account for pose uncertainty, which frequently arises in situations with ambiguous observations where a single correct pose cannot be determined. To further reduce uncertainty, we introduce a second-viewpoint module that computes complementary pose distributions, which are subsequently fused. This fusion decreases overall uncertainty and improves system efficiency. Additionally, two independent modules are included to compensate for grasping errors. The modular design allows the components to be combined for optimal performance or used individually, depending on the physical setup. 

The proposed method is evaluated in a real-world setup with three different objects, with no errors, and all modules are shown to improve efficiency. These results suggest that incorporating pose distributions with grasping pose errors is a promising direction for developing more flexible and reliable robotic production systems.  

To the best of our knowledge, this is the first framework that jointly addresses both grasping and object pose uncertainties using interchangeable modules. We believe there is ample opportunity to integrate additional modules, resulting in improved performance and flexibility. The current framework is limited to pose uncertainties in SO(2), but it could be extended to SE(3), enabling additional modules to improve the system.
\end{abstract}


\section{Introduction}

Bin-picking is the action of grasping objects randomly placed in containers. It is integral to automation as it enables the first step: object feeding. Alternatively, either manual or mechanical solutions are required, increasing the cost of a new solution. Bin-picking has therefore been extensively studied with many different solutions implemented \cite{buchholz2015bin, alonso2018current, cordeiro2022bin}.

\begin{figure}[t]
    \centering
    \includegraphics[trim=0 0 0 0,clip, width=0.93\linewidth]{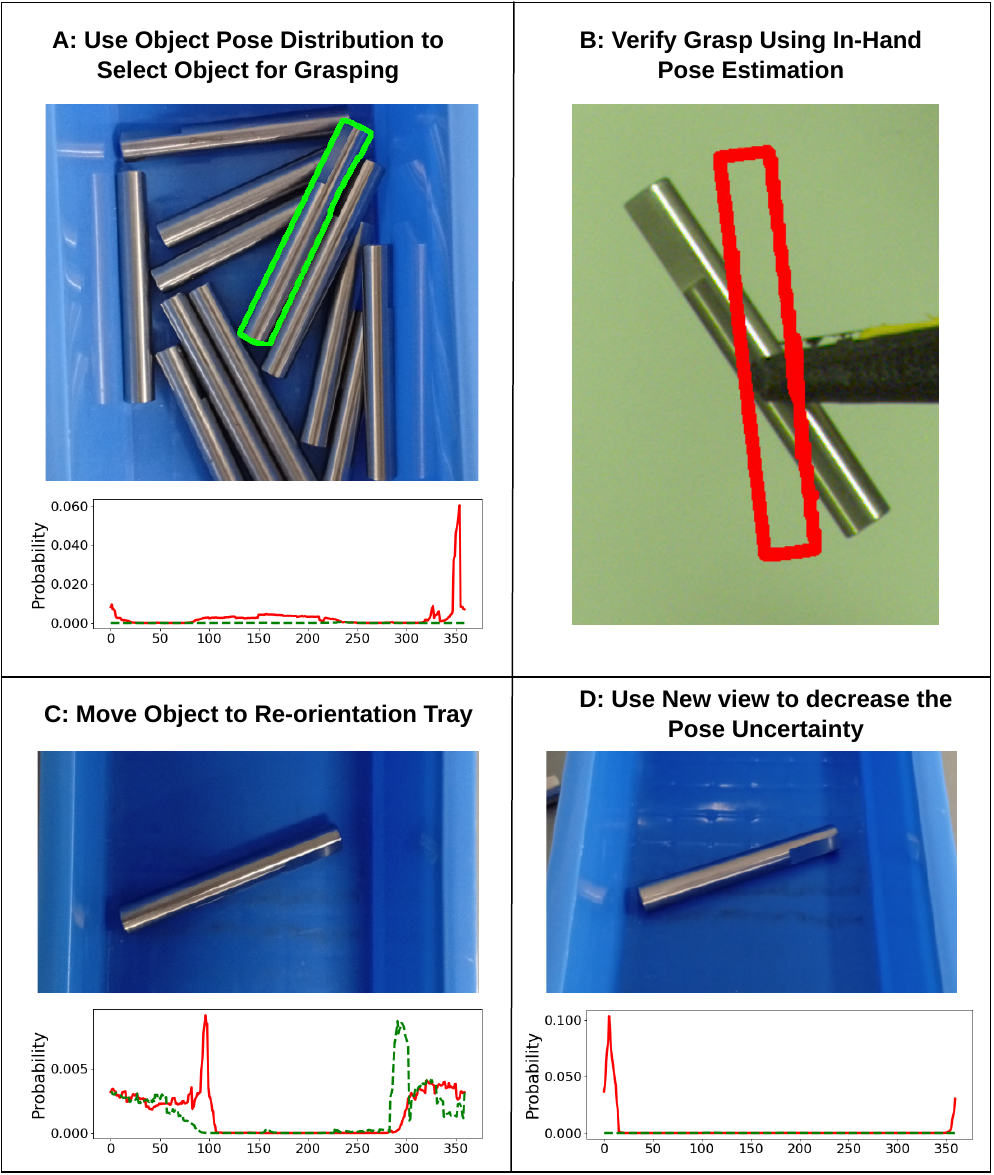}
   \caption{Example of the framework when inserting an object. \textbf{A:} First, an object with a pose distribution under the cutoff value is selected for grasping. \textit{(Object pose distribution visualized according to \cite{hagelskjaer2025good}. The red and green lines indicate the probability of revolution for the two reflections.)} \textbf{B:} The grasped object is moved in front of the in-hand camera to verify the grasp, and the expected pose (red) does not match the grasp pose. \textbf{C:} As a result, the object is moved to the re-orientation tray, and a new object pose distribution is computed. \textbf{D:} As the uncertainty is too high, a second view is obtained, and the uncertainty is re-estimated. As the new uncertainty is low and the object is in the re-orientation tray, it is inserted.}
   \label{fig:front}
     \vspace{-5mm}
\end{figure}

One key approach is to utilize visual pose estimation to identify objects for grasping. This method enables the precise positioning of the object for further manipulation, as opposed to object-less bin-picking \cite{cordeiro2022bin}. For many applications, this is a requirement for accomplishing the task and is the focus of this paper. Visual-based pose estimation also allows the system to adapt to new objects without changes in the mechanical setup. As a result, extensive research has been conducted on visual pose estimation, and the performance of current methods is very strong \cite{hodan2018bop, hodavn2020bop, sundermeyer2023bop, hodan2024bop}. 

To enable flexible production, the system should also be robust. Robustness includes the system’s ability to remain operational under varying conditions, for example, avoiding collisions, safely handling unexpected forces, and maintaining stable behavior through force‑control modes \cite{hagelskjaer2024off, schlette2020towards}. This reliability is essential to ensure the system remains online and continues operating even when disturbances occur.

But a robust feeding system should also ensure that the object is delivered in a known position. To ensure this, the system should accommodate errors in both object pose estimation and errors from the grasping process. To ensure that there are no object pose errors, the system should be able to handle uncertainty. As many objects appear ambiguous to the system due to aleatoric uncertainty, e.g., certain features are not visible from the current camera view. An example of this is shown in Fig.~\ref{fig:front}. While in many cases this ambiguity can be ignored, it can also lead to critical errors, damage to objects or robots, or the manufacture of defective products. Often, these ambiguities are resolved in subsequent actions, but this requires new sensors, mechanical solutions, and custom programming. Thus, it limits the system's flexibility and increases the setup cost.

While research in visual pose estimation has generally focused on providing single-point estimates, thereby ignoring ambiguity, there has been a growing trend toward estimating object pose distributions. Thus, methods for object pose distribution estimation have been created for providing distributions in both \textbf{SO(3)} \cite{peretroukhin2020smooth, murphy2021implicit, iversen2022ki} and full \textbf{SE(3)} \cite{haugaard2023spyropose, hsiao2024confronting, brazi2025corr2distrib}. 

However, to fully exploit the information contained in pose distributions, an appropriate decision‑making strategy must be developed. Without such a strategy, the system may simply halt when the estimated uncertainty becomes too large. 

In this paper, we introduce a modular framework designed to address this challenge. Our approach builds upon the strategy presented in \cite{hagelskjaer2025object}, but allows for adding and removing modules to improve the robustness and efficiency. To improve the efficiency, we introduce a module for obtaining a second view and combining the pose distributions to decrease the uncertainty. This allows the robot to perform much fewer object manipulations to obtain confident pose estimations.

But the strategy presented in \cite{hagelskjaer2025object} does not handle errors introduced during grasping. To handle this, we introduce two new modules, the re-orientation tray and in-hand pose verification, which are integrated into the framework. By combining the modules, the system achieves the best performance, but both modules provide robustness against grasping errors.

To the best of our knowledge, this is the first modular framework for bin-picking handling both object pose uncertainty and grasping errors. The modular structure of the framework allows the system to be integrated into a wide range of platforms, even when certain modules cannot be implemented. This modularity provides flexibility in deployment and enables the system to be tailored to different industrial settings. Additional modules can also be incorporated to further enhance performance or address application‑specific requirements.

We discuss three such modules: re‑orientation fixtures, flexible feeders, and prodding. While not implemented due to the potential decrease in system flexibility, these examples illustrate how different strategies can be combined within the proposed framework. We hope that this work will encourage further research on leveraging object pose distributions in robotics and inspire the development of new modules that expand the capabilities of flexible manipulation systems.

While our framework currently operates in \textbf{SO(2)}, leveraging only in‑plane rotations and reflections around the primary rotational axis. While this restricts the range of applicable objects, the proposed strategy is structured such that it can be extended to the full \textbf{SE(3)} space in future work. But this representation fits well with cylindrical symmetries, and as a significant number of industrial components exhibit cylindrical or near-cylindrical geometry \cite{yokokohji2019assembly, kleebergerlarge}, it has practical applications. Our focus on cylindrical symmetry also aligns with previous research in bin‑picking, such as \cite{jin2024scara}.

The main contributions presented in this paper are:


\begin{itemize}
    \item A modular framework for robust pose estimation integrating pose uncertainties and grasping errors
    \item Demonstrate use of second view for object pose uncertainty estimation in bin-picking
    \item Experiments validating each module of the strategy
\end{itemize}


\section{Related Work}
In this section, an overview of methods for robust bin-picking and methods for using object pose uncertainty in robotics is provided.

\subsection{Bin-picking}

Bin-picking is commonly divided into object-oriented and gripper-oriented approaches \cite{cordeiro2022bin}. Object-oriented methods estimate the object’s 6D pose and transfer predefined grasp poses onto it, enabling precise manipulation required for many operations \cite{yokokohji2019assembly}. But this relies on an initial correct pose estimation, which is difficult in cluttered or occluded scenes. Gripper-oriented methods instead search directly for feasible grasp configurations without relying on object pose, making them more robust in dense clutter.
In practice, the two strategies are often combined: an initial, coarse singulation grasp is used to separate objects, followed by a more precise, pose-dependent grasp for the final manipulation step \cite{cao2023two, domae2020robotic}.


One example is shown in \cite{schlette2020towards}, which combines suction gripping with a parallel gripper. First, the object is singulated from the bin using a suction gripper following a predetermined pattern. When an object is grasped, it is moved to a re-orientation table, where a pose estimation is performed. This system enabled the use of a very precise pose estimation algorithm that could not be applied to objects inside the bin. 

A method using a similar approach is presented in \cite{cao2023two}, in which a soft gripper is used to achieve both rough and precise grasping. In the bin, a deep learning-based density estimate guides the gripper to perform a rough grasp of the objects. The grasped objects are then placed in a manipulation tray, and a grasp detector determines whether to further separate them in the tray or grasp them. Similar to \cite{schlette2020towards} and \cite{cao2023two}, we also use an initial grasp to separate the objects and then perform the manipulation with the object grasped from a tray. However, we use the same algorithm to determine grasp poses for both the bin and the tray.

Similar to our method, \cite{tajima2020robust} uses pose estimation combined with predefined grasp poses to grasp the object. They then employ a tactile sensor to verify that the object has been grasped. But the sensor cannot determine whether the grasp is correct and usable for further manipulation. Instead, the object is placed in a tray for further processing. We also perform a similar grasp check, but use the gripper fingers' positions to verify the grasp. Unlike \cite{tajima2020robust}, we also verify the pose using an in-hand pose estimation system, allowing for direct manipulation of the object. We use the method described in \cite{duan2025highprecisionadaptiveselfsupervised}, which uses finger geometry to restrict object poses. However, unlike \cite{duan2025highprecisionadaptiveselfsupervised}, we only use the in-hand pose to verify that the grasp is correct. Other methods for in-hand pose estimation have also been developed \cite{choi2016using, wen2020robust}, but they do not leverage finger geometry and, as a result, yield less precise pose estimates.

Another strategy for handling objects that are not oriented correctly is to reorient them with the robot. In \cite{wan2015reorientating}, a knowledge graph is built for how to move between different poses of the object. This enables the robot to gradually rotate the object until the desired pose is obtained. We use a simplified version of this strategy to reorient objects that cannot be inserted. But in future work, this could be combined with uncertainty estimates to rotate objects in the re-orientation tray. 




\subsection{Object pose distributions in robotics}

The method most similar to ours is the bin-picking pipeline presented in \cite{hagelskjaer2025object}. Here,  \textbf{SO(2)} rotational pose distributions are estimated for objects placed in a bin. A grasping and re-orientation strategy is then created to ensure that only objects with unambiguous pose estimates are grasped. The bin-picking strategy of \cite{hagelskjaer2025object} is further elaborated in Sec.~\ref{sec:method:orig}. We extend this pipeline into a modular framework for robust and efficient bin-picking.

One approach to handling uncertainty in robotics is to guide the grasping process \cite{naik2025robotic}. Here, the task execution is simulated by testing different grasp poses with the measured object pose uncertainty. This provides a measure of robustness for each possible grasp pose, allowing the most robust pose to be selected or more data to be gathered. A different approach is presented in \cite{johns2016deep} where a physics simulation is used to train a neural network to predict the probability of grasping success. At run-time, the best grasp pose is then the region with the greatest possibility of success. In this paper, we do not model grasping uncertainty, as our experiments show that for the objects, errors in object pose estimates do not significantly affect grasping success. Instead, grasps fail due to non-prehensile manipulation during the process: other objects are moved, thereby moving the intended grasping object. The approach presented in \cite{johns2016deep} could be implemented in our method to select grasp poses. But the simulation requirement makes the method less flexible, and it is therefore not used. 

Another use of object pose distributions is active perception. In \cite{jin2025se}, it is combined with grasp planning and demonstrates improved performance over the baseline method. The active perception is used to predict the best viewpoint by sampling around the object. However, it is limited to a fixed object distance. The robot then moves the sensor to this position to reduce the uncertainty. Active vision is also used in \cite{naik2025robotic}, where the pose distribution is refined using multiple cameras and views. 

We also use multiple views to refine the object pose distribution, but instead of actively selecting the best next pose, we fix the second view at a predefined position. This is because, unlike \cite{jin2025se} and \cite{naik2025robotic}, our objects are placed inside a bin. Many views would thus be obscured, and as the robot cannot move freely in the space, many poses cannot be reached. Instead, we manually define the second view that gives the best view into the bin.

\section{Method}
\label{sec:method}



The method presented in this paper is a modular strategy for flexible and robust bin-picking, based on the system presented in \cite{hagelskjaer2025object}, to which we add modules that ensure robustness and increase efficiency. The system is flexible as it allows adding or removing modules depending on the current workcell and the desired operation. The added modules are a re-orientation tray, a second view, and an in-hand pose estimation. We also discuss three additional modules, fixture check, vibratory feeder, and non-prehensile manipulation, which are not currently added to the system, but could be in future work. In the following section, we elaborate on each module and discuss its feasibility, and finally present the complete system.

\subsection{Original method}

\label{sec:method:orig}

\begin{figure}[t]
    \centering
    \includegraphics[angle=0, width=.60\linewidth]{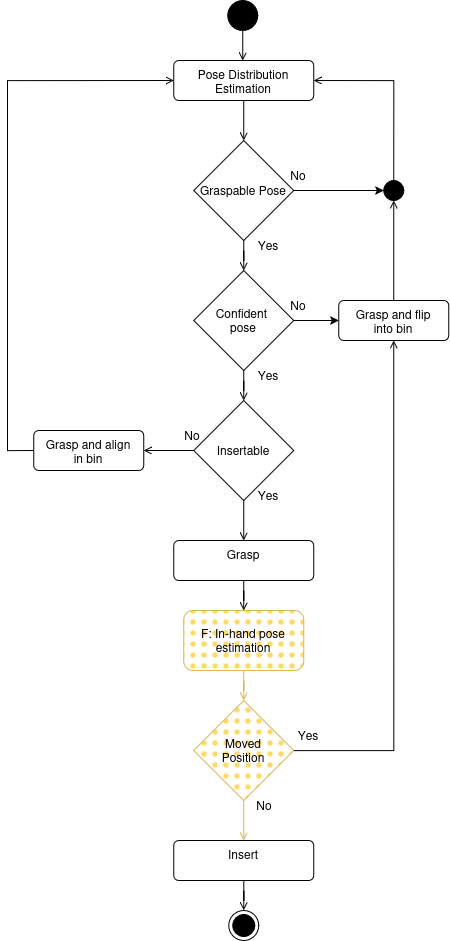}
    \caption{Activity diagram of the bin-picking process with an added in-hand pose verification module. The original method \cite{hagelskjaer2025object} is shown in clear white, while the in-hand pose verification module is shown in yellow dots.}
    \label{fig:strat}
\end{figure}

In \cite{hagelskjaer2025object}, a simple strategy was developed. First, a precise object pose is found using \cite{hagelskjaer2024keymatchnet}, and the \textbf{SO(2)} rotational pose distribution is then computed. A decision tree selects the appropriate action based on the estimated distribution of object poses. While the method successfully reoriented the object into a graspable pose without pose ambiguity, it has two main drawbacks. Firstly, it is time‑consuming because the object is re‑oriented inside the bin rather than on a stable surface. Thus, it can easily move during this process. Secondly, the object feeding is unreliable. Because the object is grasped from a cluttered bin, it may move during grasping, leading to erroneous grasps. These errors can then result in failed insertions.



\subsection{In-hand pose verification} 

The first module introduced is the in‑hand pose verification. This module is applied after the object has been grasped to verify that it has not moved during the grasp. We use the same pose‑estimation method as in \cite{hagelskjaer2025good, duan2025highprecisionadaptiveselfsupervised}, but we do not use the estimated pose directly for the insertion. Instead, we compare the in‑hand object pose with the expected pose, similar to the approach in \cite{hagelskjaer2025good}, Eq.~2. The reason we do not use the in-hand pose estimate for insertion is twofold: firstly, online planning of the new pose could result in unexpected behavior from the robot. Secondly, and more importantly, the actual pose cannot be determined from the in-hand pose estimate as it is difficult to inspect the object from all angles \cite{duan2025highprecisionadaptiveselfsupervised}. However, we can compare the object outline with the expected position, and if they match, we assume the object was grasped correctly. Therefore, if the estimated in‑hand pose does not match the expected pose, we simply reject it and retry the grasp. This ensures that no incorrect insertions are performed. The original strategy from \cite{hagelskjaer2025object}, with the added in‑hand pose verification module, is shown in Fig.~\ref{fig:strat}.






%

\subsection{Re-orientation Tray} 

Another strategy to remove errors introduced during grasping is a re-orientation tray. When the object is grasped in the bin, it is moved to the re-orientation tray, and a new pose distribution estimate is computed. When the object is then grasped from the tray, it lies flat on the bottom, and no other objects influence it. 

This singulation serves two purposes. Firstly, it makes the process more efficient because the reorientation and flipping actions are unaffected by other objects. Thus, the intended robotic action is more likely to succeed as the object does not get stuck on other objects. The second advantage is that objects grasped from the re-orientation tray are much more precise. This means the object can be grasped without error. Thus, without any external sensors or specialized hardware, a robust pose estimation system can be created. The system can, of course, also be combined with the in-hand pose verification to improve efficiency.



\subsection{Fixture verification} 

An alternative strategy to ensure the object is correctly grasped is to use a fixture for verification. This fixture is used after grasping but before the final insertion. If the object was grasped correctly, it will fit in the fixture; otherwise, it will not. This approach has shown strong results in \cite{duan2025highprecisionadaptiveselfsupervised}, but this module has not been added to our system because it generally requires a novel fixture for each new object, reducing flexibility. But, for some applications, fixtures are already available, or adding one is not a great cost.

\subsection{Second View}

For some object viewpoints, the system cannot reliably determine the correct orientation. While the strategy presented in \cite{hagelskjaer2025object} ensures that objects are reoriented until their pose distribution falls below a specified cutoff, the reorientation step can be time-consuming.

An alternative strategy is to acquire additional camera views to reduce pose uncertainty. In our system, the primary view used to estimate object pose distributions is a top-down view of the bin. We define a secondary view angled 30 degrees from the front. This view is chosen based on the bin geometry, which includes a frontal opening. A static second view is used to avoid introducing errors in the robot planning algorithm. The primary and secondary views are shown at the bottom of Fig.~\ref{fig:front}.

To compute the pose distributions from the second image, the transformation between the two camera positions is first determined as follows:
\begin{equation}
\ _{cam2} T^{cam1} = ( \ _{base}T^{cam2})^{-1} \ _{base}T^{cam1}
\end{equation}
where $_{base}T^{cam1}$ and $_{base}T^{cam2}$ are the transforms from robot base to camera position, for the first and second view, respectively. Finally, we transform the object pose from the first camera, $_{cam1}T^{obj}$, into the second camera's view, $_{cam2}T^{obj}$:
\begin{equation}
\ _{cam2}T^{obj} = \ _{cam2} T^{cam1} \ _{cam1}T^{obj}
\end{equation}



With the two view-points we now have object pose distribution estimates for both images $p( \theta  ~ | ~ I_1 )$ and $p(\theta  ~ | ~ I_2 )$. Where $I_1$ and $I_2$ are the first and second images respectively, and $\theta$ is the object pose which we are estimating. By assuming the images are independent given the object poses: 
\begin{equation}
I_1 \perp I_2 ~ | ~ \theta
\end{equation}
we use the Product of Likelihoods \cite{monahan1992proper} to combine the distributions as follows:
\begin{equation}
p(\theta ~ | ~ I_1,I_2) = p(\theta ~ | ~ I_1) ~ p(\theta ~ | ~ I_2)
\end{equation}
We then use the combined distribution, $p(\theta ~ | ~ I_1,I_2)$ to determine if the object pose distribution estimate is within the cut-off value.





\subsection{Vibratory Feeder}


A different strategy to reduce the aleatoric uncertainty of the objects is to introduce movement into the system. Vibratory feeders have demonstrated good performance \cite{malik2019advances}. However, these solutions require emptying the bin into the feeder and refilling it after each grasping cycle, which is highly inefficient for small batch sizes. In addition, vibratory systems often require careful tuning of the vibration parameters for each object type, reducing the overall flexibility of the setup. It is, therefore, not included in our current framework. Nevertheless, for certain applications with stable part geometries and larger throughput requirements, such solutions may still be acceptable.

\subsection{Non-prehensile Manipulation}

Another method for reorienting the objects is to use the gripper for non-prehensile manipulation. If used in the bin, it enables reorientation of multiple objects, similar to a vibratory feeder \cite{cao2023two}. While this does not require new hardware for the setup, the fine-tuning of the task can be quite cumbersome \cite{dogar2012planning, lynch2003control, serra2016robot}. If too little force is applied, the objects won't move, and if too much force is applied, they might spill out of the bin. It could also be applied in the re-orientation tray \cite{wan2015reorientating}. In this scenario, a single object could be reoriented according to a predefined strategy to obtain more certain pose estimates or to allow insertion. We do not use either application, but this could be explored in further work.


\begin{figure}
    \centering
    \includegraphics[width=0.89\linewidth]{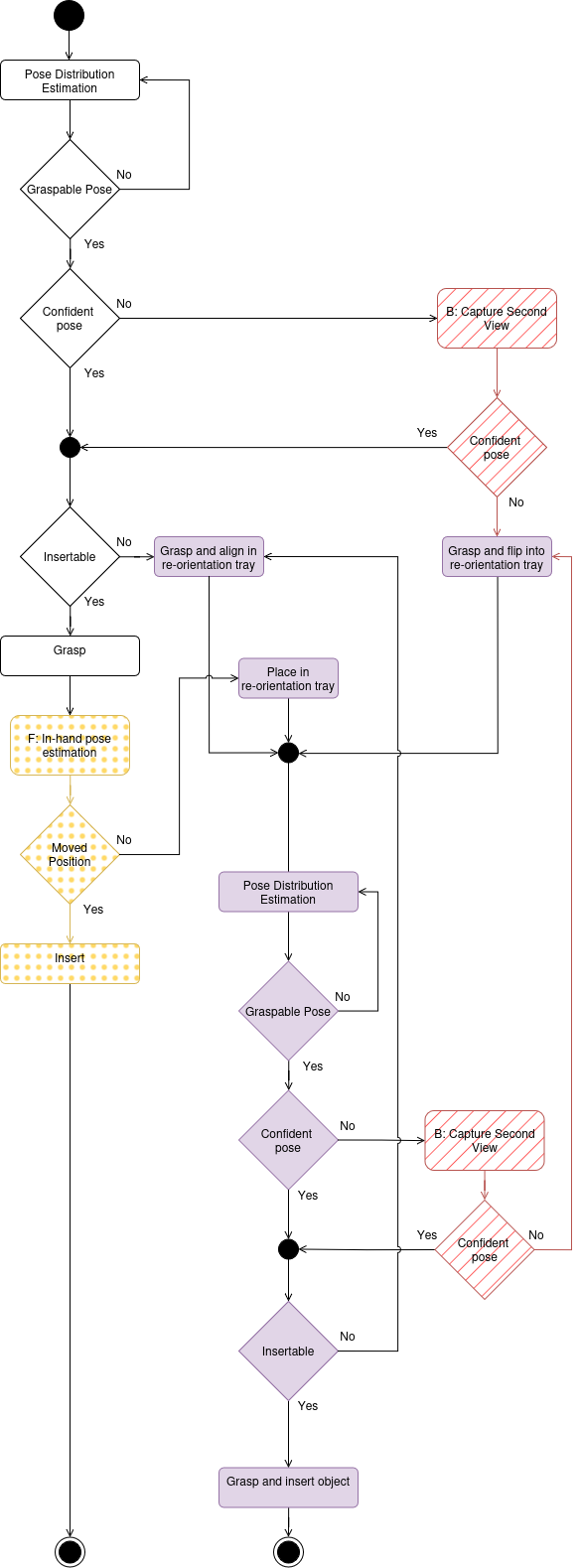}
    \caption{The framework for the proposed bin-picking strategy. White: Original system. Striped Red: Second view. Dotted Yellow: In-hand pose verification. Purple: Use of re-orientation tray. If either the In-hand pose verification or Second view is omitted the flow simply follows the "No" result of the decisions in each of these modules.}
    \label{fig:activity}
\end{figure}

\subsection{Modular Framework}

The overall framework combines the strategy presented in \cite{hagelskjaer2025object} with In-hand pose verification, the Re-orientation Tray, and the second view to estimate the object pose distribution. The system starts by estimating the object pose distribution in the bin. If no poses are found, it retries the pose estimation. If objects are found but the pose is uncertain, the second view is used to reduce the uncertainty. If no specific poses are found, the best grasp pose, as per \cite{hagelskjaer2024off}, is used, and the object is flipped into the re-orientation tray. If an object is certain but no insertion pose can be calculated, it is moved to the re-orientation tray and rotated into an insertion pose. If a certain, insertable pose is available, it is grasped and moved to the In-hand pose verification. If the pose matches, the object is inserted; otherwise, it is placed in the re-orientation tray.

If the object is moved to the re-orientation tray, regardless of the reason, the process now restarts. Here, pose estimation and grasping are performed on the re-orientation tray. The only difference is that the In-hand pose verification system is not used, as the object is grasped from a flat surface. The overall strategy with all modules is shown in Fig.~\ref{fig:activity}.




\section{Experiments}


\begin{figure}[t]
    \centering
        \includegraphics[trim=0 0 0 0,clip, width=0.95\linewidth]{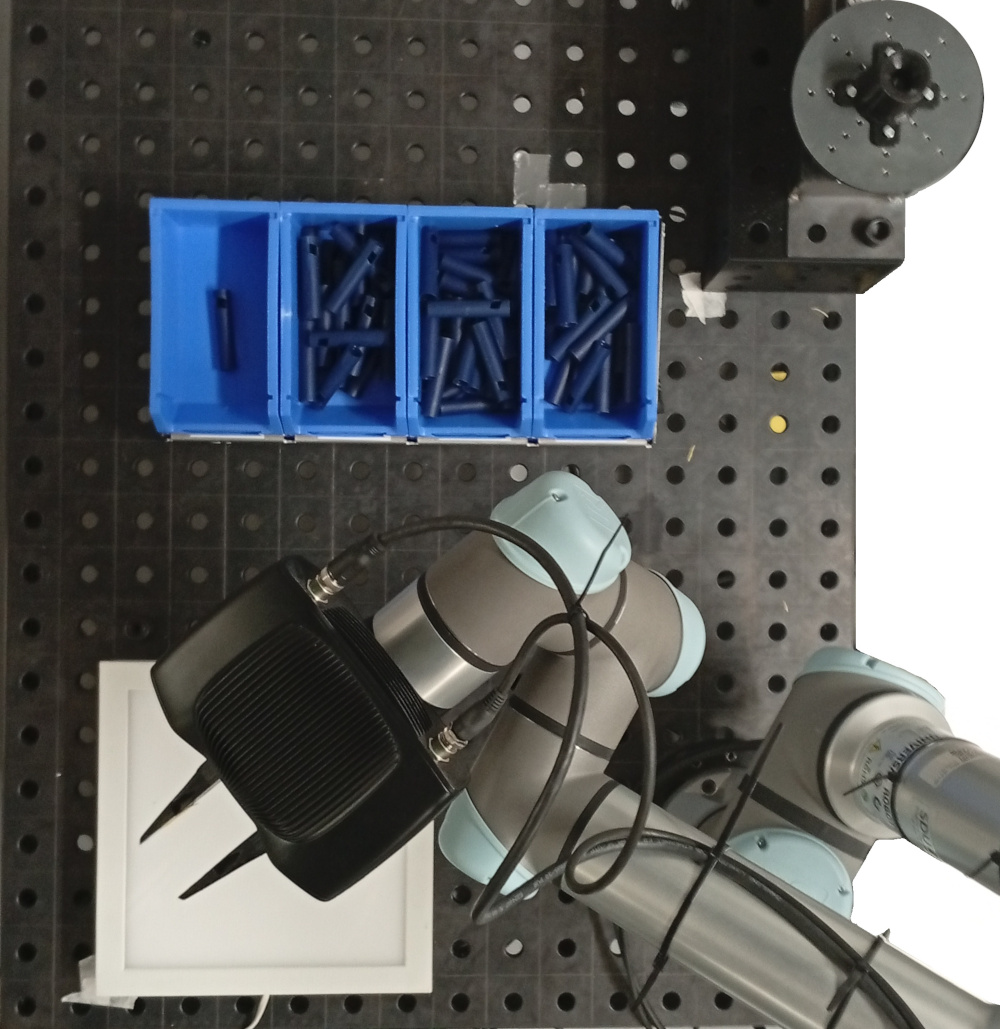}
        
        \caption{The workcell used for experiments. The overhead camera for in-hand inspection is placed directly above the white backlight. The leftmost bin is used for re-orientation tray. Top right is the re-orientation fixture used in \cite{duan2025highprecisionadaptiveselfsupervised}.}
    \label{fig:workcell}
\end{figure}

\begin{figure}[t]
\centering
    \includegraphics[trim={0cm 0.2cm 0cm 0.20cm},clip,width=.99\linewidth]{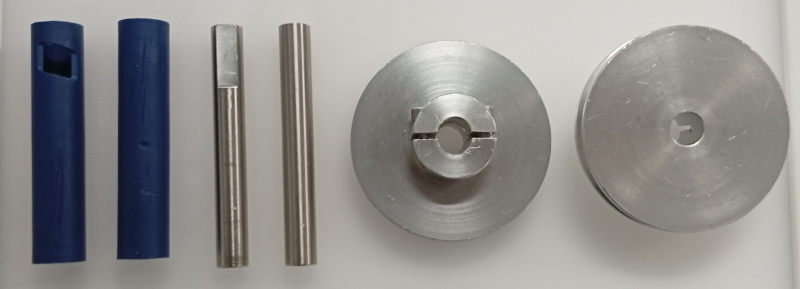}
   \caption{The objects used in the experiments shown from both sides. For the Novo object (left) the square recess is used to obtain both reflection and revolution. For WRS-08 (middle) the reflection and revolution is found by the flat surface. For WRS-11 (right) the reflection is found by the presence or lack of the clamp, but the revolution can only be determined from the clamp.}
   \label{fig:objects}
\end{figure}

The method is tested on the bin-picking workcell presented in \cite{hagelskjaer2025good}. The workcell contains a rack of bins with objects, a UR-5 robot with a Zivid2 3D sensor and a Robotiq HAND-E gripper with 3D-printed fingers, an overhead RGB camera for in-hand inspection, and a fixture for verifying the pose. The workcell is shown in Fig.~\ref{fig:workcell}.

We compare our method with the method presented in \cite{hagelskjaer2025object} by using different combinations of the modules presented in Sec.~\ref{sec:method}. The main experiments are performed using the same object as in \cite{hagelskjaer2025object}. The objects used for experiments are shown in Fig.~\ref{fig:objects}. We measure the percentage of successful insertions and the average number of grasps required per insertion. For each test, we run until 40 successful insertions have been performed. The results are shown in Table \ref{tab:res}. 

The results indicate that the best overall performance is achieved by using all three modules. It is also seen that the greatest reduction in the number of grasps is obtained with the re-orientation tray, followed by the new view. It is also evident that both In-hand pose verification and the Re-orientation tray ensure a 100\% success rate for insertions. Thus, robust bin-picking can be achieved with either module, but the best overall performance is obtained by integrating both and also utilizing the new view.


\begin{table}[t]
    \vspace{1.5mm}
    \label{tab:res}
    \caption{Resulting success rate and number of grasps for the framework with different modules added. }
    \begin{tabular}{lcc}
    Strategy                    & Success rate & Grasps per insertion \\ \hline
    Original (O)                &  89            &    2.5                  \\
    O + In-hand Pose (IHP)      &  100           &    2.8                  \\
    O + Reorientation Tray (RT) &  100           &  2.25                    \\
    O + RT + New View (NV)      &  100           &  2.05                   \\
    O + RT + NV + IHP           &  100           &  1.91                   
    \end{tabular}
\end{table}

\subsection{Experiments on WRS challenge}

We also test the effectiveness of the developed method on two objects from the World Robot Summit Assembly Challenge, \cite{yokokohji2019assembly}. Specifically, we consider the kitting part, where objects should be grasped from a bin and placed in a kitting tray. We selected these objects because of their low success rate during the competition. The pulley, WRS-11, was the most difficult, with an average success rate of 1.6\%, and the drive shaft, WRS-08, had an average success rate of 10.0\%. 

Our system is based on the workcell presented in \cite{hagelskjaer2024off}, which successfully performed the kitting task. However, because the only requirement for completing the kitting challenge was to place the objects inside the tray, it did not require them to be placed in a specific orientation, but in the subsequent assembly step, the correct reflection was provided. We therefore extend the task to require the correct reflection of the objects as well.

Using the created framework, we perform 10 kittings of each of the two test objects placed inside the bin. The system correctly performs reflection-discriminative kitting for each object, achieving a 100\% success rate. On average, it required 2.2 and 3.3 grasps per object for WRS-11 and WRS-08, respectively. It is possible that the metallic surface increased the pose uncertainty relative to the Novo object, resulting in more flip operations. While preliminary, these results show that the method is extendable to other industrial objects and tasks.

\section{Conclusion}

In this paper, we present a modular framework for robust bin-picking using object pose distributions. We present three modules for the framework: in-hand pose estimation, re-orientation tray, and a second-view distribution estimation. The framework is tested on a real use case and demonstrates robust bin-picking. The modular framework allows for the integration of additional modules to either increase efficiency or flexibility. 

Future work includes extending the framework to \textbf{SE(3)}, which would both enable seamless integration into a full 6‑DoF setting and enhance its expressiveness, for instance, by explicitly modeling grasping uncertainty. Further improvements may be achieved by incorporating additional sensing modalities, such as RGB information or tactile feedback, in the estimation of the object‑pose distribution. The in-hand pose estimation could also be changed from a single point estimate to an object pose distribution, such as \cite{haugaard2023spyropose}.

\bibliographystyle{IEEEtran}
\bibliography{egbib}

\end{document}